\documentclass{Interspeech2024}
\usepackage{multirow}
\usepackage{graphicx}
\usepackage{url}
\usepackage{multicol}
\usepackage{color}
\usepackage{cancel}
\usepackage{svg}

\title{MS-HuBERT: Mitigating Pre-training and Inference Mismatch in Masked Language Modelling methods for learning Speech Representations}

\name[affiliation={1}]{Hemant}{Yadav}
\name[affiliation={2}]{Sunayana}{Sitaram}
\name[affiliation={1}]{Rajiv Ratn}{Shah}

\address{
  $^1$IIIT Delhi, India \\
  $^2$Microsoft Research Banglore, India
  }
  
\email{\{hemantya,rajivratn\}@iiitd.ac.in, sunayana.sitaram@microsoft.com}

\interspeechfinaltrue 


\begin{document}

\maketitle

\begin{abstract}
In recent years, self-supervised pre-training methods have gained significant traction in learning and encoding high-level information from speech data. Among these methods, HuBERT has demonstrated state-of-the-art performance in automatic speech recognition (ASR). However, HuBERT's performance lags behind data2vec due to disparities in pre-training strategies. In this paper, we propose MulticlusterSwap-HuBERT (MS-HuBERT), which integrates a Swap method to address pre-training and inference mismatch observed in HuBERT and incorporates Multicluster masked prediction loss for more effective utilization of the models capacity. MS-HuBERT, an end-to-end self-supervised pre-training method for robust speech representation learning, beat vanilla HuBERT on the ASR Librispeech benchmark by a large margin.  Additionally, we demonstrate that the embeddings obtained during pre-training encode essential information for improving ASR performance. The model is available to use in SUPERB repository \footnote{https://github.com/s3prl/s3prl}. 
\end{abstract}

\section{Introduction}
\label{intro}
In the recent years, there has been a significant interest in studying self-supervised pre-training methods to learn/encode high level information present in the speech data \cite{baevski2020wav2vec2.0, hsu2021hubert,brown2020languagegpt3, chung2019unsupervised,oord2018representationcpc}. These SSL methods utilize the input data itself to learn to encode useful information, with the choice of pretext task playing a pivotal role in the encoded information. The most popular pretext task used is masked predictive coding (MPC) \cite{schneider2019wav2vec, baevski2020wav2vec2.0,chung2021w2vbert, discretebert, chen2022wavlm}. 
HuBERT \cite{hsu2021hubert} is one such model that popularized the masked language modelling (MLM) technique to learn high-level speech representations from raw audio by achieving SOTA on the automatic speech recognition (ASR) task. The underlying concept of HuBERT revolves around iterative pre-training: starting with a raw audio/pseudo-label pair (x/y), the model undergoes successive training iterations where the trained model updates the pseudo-labels, iteratively refining its representations until a predefined stopping criterion is reached. I However, despite its success, HuBERT falls short compared to data2vec \cite{baevski2022data2vec} in ASR performance for two primary reasons: firstly, during pre-training, data2vec accesses the full context to generate continuous labels, which are updated after each gradient update step, as opposed to the fixed discrete labels utilized in HuBERT for the each iteration; and secondly, the output is averaged from multiple layers for loss calculation. 

To bridge this gap, we propose two modifications to the HuBERT framework. Firstly, we introduce the "Swap" method to enable full context access during pre-training, thus addressing the pre-training and inference mismatch observed in HuBERT and other MLM-based methods by using both the masked and unmasked views during pre-training. Swap is motivated by a simple idea, used heavily in the field of computer vision \cite{chen2020simplesimclr, he2020momentummoco, grill2020bootstrapbyol}. Where two augmented views of the input are used to learn a high level representation.  Given $e$ layers in a encoder, it is a general practice to add a similarity loss on the output embeddings of the encoder after each layer, as seen in works using U-net type architectures \cite{ronneberger2015unet1,riahi2023singleunetconcatenate,macartney2018improvedunetconcatenate2}. In contrast, we propose a Swap method which swaps the output embeddings, at certain indices, after each layer of the encoder between the masked and unmasked view of the input. This is motivated by the simple fact, that the learned model is expected to generate exactly the same output regardless of the two views. 

Secondly, inspired by the work of Yadav et al.  \cite{yadav2023analysing}, we adopt a Multicluster masked prediction loss (MPL) approach. Using multiple cluster centers, also called multiple resolutions, has been investigated by \cite{shi2023explorationMRHUBERT, shi2023MRHuBERT, yadav2023analysing}. 
In \cite{shi2023MRHuBERT}, the author introduces down-sampling and up-sampling modules within the transformer encoder after each layer to facilitate learning features at multiple resolutions. On the other hand, \cite{shi2023explorationMRHUBERT} explores parallel and hierarchical variations of HuBERT with findings indicating the superiority of the hierarchical approach. This involves training multiple models, each model adds almost same parameters as the original HuBERT, at various resolutions using CNN as a down-sampling module. Lastly \cite{yadav2023analysing} uses the fact that MPL is applied at multiple layers of encoder at different resolutions. This method does not introduce any additional parameters to the original HuBERT model, except the linear layers used for loss calculation which are discarded after the pre-training. In this work, we adopt this approach and modify it for our use case for loss calculation.

These changes align HuBERT more closely towards data2vec, primarily differing in their loss functions for pre-training and other minor changes. The goal is study how much HuBERT can be improved, with these changes, on the ASR task.

\begin{figure*}[ht]
\centering
\includegraphics[trim=13cm 6.25cm 14cm 6.8cm, clip,width=2\columnwidth]{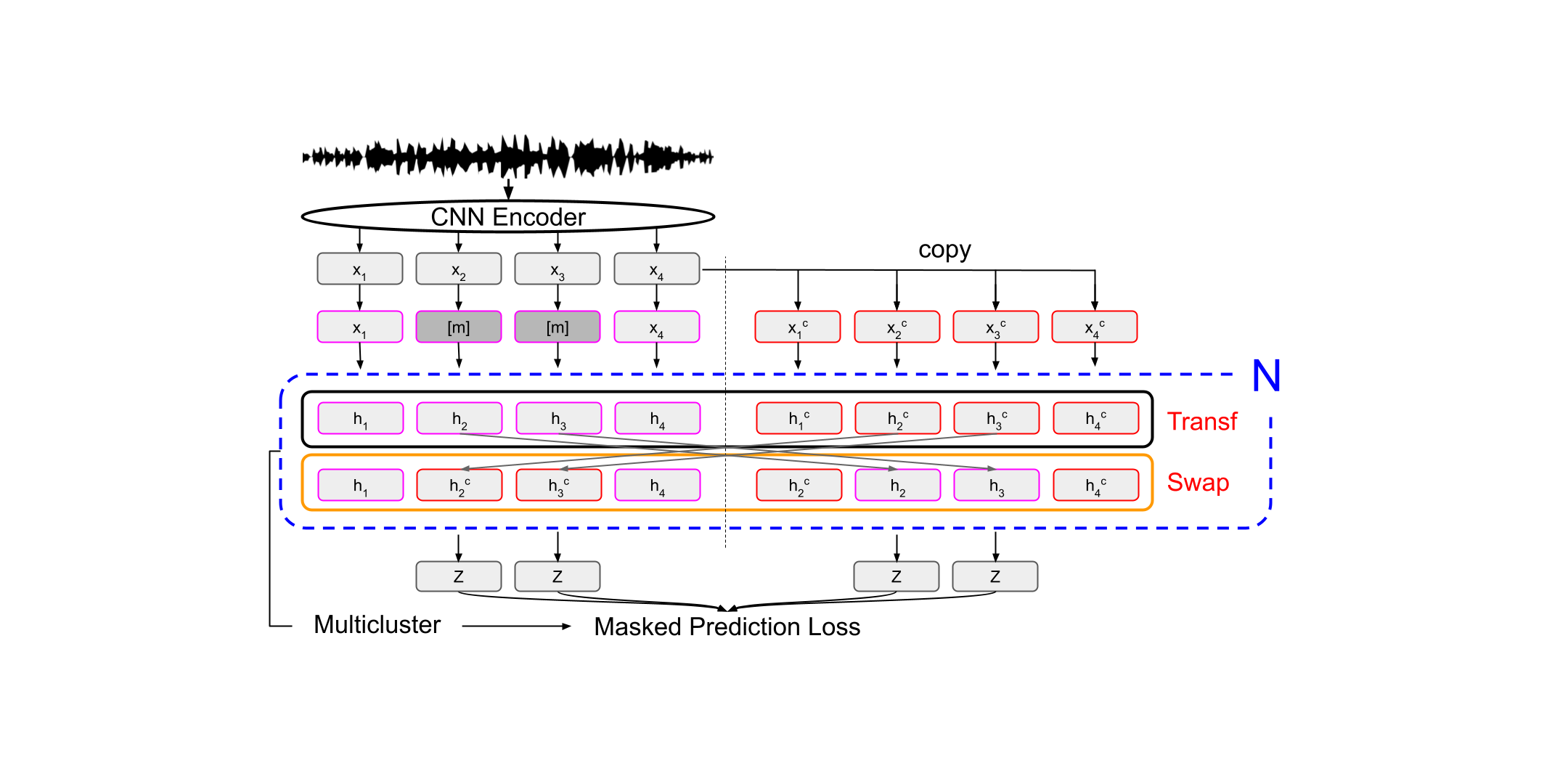}
\caption{Proposed MS-HuBERT approach, an end-to-end self supervised pre-training method to learn robust speech representations. The input raw audio is passed to a CNN encoder. Two copies of the output is created i.e., masked and unmasked. Which is passed through the Swap modified 2nd encoder. Multicluster Masked prediction loss is calculated, masked indices only, on the output embeddings from different blocks of the modified 2nd encoder.}
\label{fig:MS-HuBERT method}
\end{figure*}

Based on these observations. In this work, we propose MulticlusterSwap-HuBERT (MS-HuBERT) method, which incorporates (i) the Swap method to address the pre-training and inference mismatch issue, as the [MASK] symbol never appears during the inference, and (ii) the  Multicluster MPL similar to \cite{yadav2023analysing}. Our contributions are as follows:

\begin{enumerate}
    \item We propose MS-HuBERT, an end-to-end self supervised pre-training method to learn robust speech representations. It combines the Swap method and Multicluster MPL with HuBERT as shown in Figure \ref{fig:MS-HuBERT method}. 
    
    \item We show that MS-HuBERT outperforms the original HuBERT on the ASR Librispeech benchmark with a big margin. And matches the performance of data2vec in high-resource setting. 

    \item We showcase that the embeddings acquired during pre-training encode crucial information essential for addressing the ASR task. Thus utilizes the model capacity very effectively.
    
\end{enumerate}

\section{Method}
\label{section:method}
\subsection{Background}
HuBERT is an iterative pre-training SSL method comprising of two encoders based on CNN (1st) and transformer (2nd), in that order, architectures. The CNN encoder serves the dual purpose of down-sampling the input data. The resulting output is passed, denoted as $U$, to the transformer encoder and its output is used for loss calculation. During the pre-training stage, raw audio is passed to the CNN encoder and approximately $50\%$ of the output is masked, using the masking token $[M]$ and is subsequently passed to the transformer encoder. The network is then trained to optimize to output a discrete target sequence by minimizing the masked prediction loss. The complete details can be found in the original paper \cite{hsu2021hubert}.

\subsection{MS-HuBERT}
MS-HuBERT augments HuBERT model in two ways (i) the Swap method and (ii) the Multicluster MPL as shown in Figure \ref{fig:MS-HuBERT method}. Swap method is introduced to address the pre-training and inference mismatch phase in HuBERT i.e, during inference the model does not use masking. Swap method modifies the 2nd encoder of HuBERT, such that the updated model now encounters, two views of the input, both masked and unmasked inputs during pre-training. Lastly, our proposed method uses modified Multicluster MPL as proposed by \cite{yadav2023analysing}, because of its enhanced model capacity utilization in learning features suitable for the ASR task. These changes aim to improve the ASR performance, as shown in the Table \ref{table:ASR finetuning}.

\subsubsection{Swap}

Given a raw audio as an input, of batch of size 1, to the 1st encoder (CNN), its output is denoted as $X = x_1, x_2, ...,x_{t-1},x_t$, where $t$ represents the total number of output tokens. 
Two views of $X$ are created: (i) masked view, where on average, around 50\% of these tokens are masked, meaning that half of these tokens are replaced with the $[m]$ token, resulting in an updated output $X^m = x_1, [m], ...,[m],x_t$ (view 1) and (ii) unmasked view, a duplicate of the original $X$, denoted as $X^c = x_1^c, x_2^c, ...,x_{t-1}^c,x_t^c$ (view 2). These two views are combined, to form a batch of size 2, and is passed to the 2nd encoder.

The second encoder has $N$ layers, each composed of a transformer layer followed by a Swap layer as shown in Figure \ref{fig:MS-HuBERT method}. The transformer layer is exactly similar to the original HuBERT method. The proposed Swap method's function is to swap the outputs, at the masked indices, of the transformer layer between the two views. This updated output serves as input to the next block of encoder layer, and the process repeats till the last layer. For example, the output of the transformer layer is $H_m = h_1, h_2, ..., h_{t-1}, h_t$ and $H^c = h_1^c, h_2^c, ..., h_{t-1}^c, h_t^c$ for the masked and unmasked input respectively. The outputs at the masked indices are now swapped using the swap method i.e., the updated output are $H^m = h_1, h_2^c, ..., h_{t-1}^c, h_t$ and $H^c = h_1^c, h_2, ..., h_{t-1},h_t^c$ for the masked and unmasked input, respectively. 

It's important to note that there is no associated loss with the "Swap" layer. This technique indirectly encourages the model to output the same embeddings irrespective of the masked and unmasked view.

\subsubsection{Multicluster MPL} The Multicluster MPL, inspired from \cite{yadav2023analysing}, involves the computation of masked prediction loss (MPL) across multiple layers of the transformer encoder, using multiple set of cluster centers as labels. These encoder layers are selected equidistant in between the last layer and one intermediate layer. 
For instance, consider a scenario with three sets of labels as $(500, 250, 100)$, where the last layer index is 12 and the intermediate layer index is 8, the multiple layers are $(12,10,8)$. 

The Multicluster MPL is then formulated as the summation of MPL over $a$, where $a={(12,500), (10,250), (8,100)}$ is a dictionary of which label set to use with which transformer encoder layer \footnote{In the original paper $a$ would be calculated in reverse order i.e.,  ${(8,500), (10,250), (12,100)}$.}. MPL is computed over the masked indices only, as depicted in Figure \ref{fig:MS-HuBERT method}. Furthermore, given the GPU memory constraints, we randomly drop $d$ items from the  dictionary $a$ for every forward pass.
\begin{center}
    $Multicluster \ loss = \sum_{a}^{} (MPL)$.     
\end{center}

\section{Experimental Details}
\label{section:experimentaldetails}

For all the experiments, similar to the HuBERT base model configuration \cite{hsu2021hubert}, the MS-HuBERT model comprises a CNN encoder and 12 encoder transformer layers consisting of 768-dimensional hidden states and 8 attention heads. There is no large model used for training or comparison purposes.  

\noindent \textbf{Datasets}: The ASR Librispeech benchmark dataset \cite{panayotov2015librispeech}, which is derived from the LibriVox project, is used for pre-training and supervised finetuning purposes. It has 3 splits (i) Training, comprising train-clean-100, train-clean-360, and
train-other-500, (ii) Development including dev-other and dev-clean, and (iii) Testing consists test-other and test-clean. Each data instance comprises an audio and its corresponding transcript. For pre-training MS-HuBERT, we use only the raw audios from the combined training split resulting in a total 960 hours audios. For supervised fine-tuning, three sets of Libri-Light \cite{librilight}: 1 hour, 10 hour, 100 hour and the full Librispeech 960 hours dataset is used.

\noindent \textbf{pseudo-labels}: Six sets of pseudo-labels with varying numbers of clusters/resolutions are generated using first iteration HuBERT \cite{hsu2021hubert}. Initially, a K-means model with 1000 cluster centers is trained using latent features extracted from the 6th layer of the first iteration HuBERT base. Subsequently, another K-means model with 500 cluster centers is trained using the 1000 cluster centers as features obtained in the prior step. This process is iteratively repeated four times to train four more K-means models with 250, 125, 50, and 25 cluster centers (in that order) utilizing the cluster centers extracted from the previous step. This results in a total of 6 set of pseudo labels used to calculate the Multicluster MPL.

\noindent \textbf{Pre-training}: Unlike HuBERT, MS-HuBERT base incorporates 6 classification heads instead of just 1. This is because of the Multicluster MPL. This results in a total parameter count of 96.01 million, representing an increment of around 1.25 million parameters compared to HuBERT. MS-HuBERT is trained for 400,000 iterations on 32 GPUs with a batch size of at most 87.5 seconds of audio per GPU. The best model checkpoint is determined using the dev-other subset. Pre-trained models and training configurations will be made available after the acceptance. 

Given the memory constraints and to avoid the out-of-memory error, we randomly drop 2 clusters, and their respective layer indices, in each gradient update step. Furthermore, the intermediate layer index is chosen using the formula : $0.25*12$, where $12$ is the number of transformer encoder layers. 


\noindent \textbf{Supervised Fine-tuning and inference}: We follow the Wav2Vec 2.0 \cite{baevski2020wav2vec2.0}  strategy to fine-tune MS-HuBERT to minimize the Connectionist Temporal Classification \cite{graves2006connectionist} loss using 8 GPUs.
The total batch size is of 200 seconds of audio per GPU and the best model checkpoint is determined by the lowest Word Error Rate (WER) achieved on the dev-other split. 
For inference, 4-gram language model (LM) is used with a beam width of 500 for dev-other, dev-clean and 1500 for test-clean and test-other. 
We do a conservative hyper-parameter search for the 1 hour and 10 hour splits and fixed hyper-parameter are used for the 100 and 960 hours training splits during fine-tuning. The inference hyper-parameters are searched with Ax,
a Bayesian optimization toolkit \footnote{https://github.com/facebook/Ax} with a beam-width of 500 using 32 trials.


\section{Results}
\label{section:results}

\renewcommand{\arraystretch}{1.2}
\begin{table}[t]
\caption{ASR Librispeech benchmark finetuning results using a 4-gram language model. wav2vec 2.0 and data2vec are not a direct comparison to MS-HuBERT and is shown only such that the reader has a broader picture. The readers should ignore these two models until the Discussion section \ref{section:discussion}.}
\centering
    \scalebox{0.8}{
    \begin{tabular}{l|c|c|c|c}
    \hline
    
      Method & dev-clean  & dev-other & test-clean  & test-other  \\
        \hline
        \hline
        \multicolumn{5}{c}{\multirow{1}{*}{\textbf{1hr}}} \\   
        \hline
        wav2vec 2.0 \cite{baevski2020wav2vec2.0} & 5.0 & 10.8 & 5.5 & 11.3 \\
        HuBERT \cite{hsu2021hubert}  & 5.6 & 10.9 & 6.1 & 11.3  \\
        WavLM \cite{chen2022wavlm} & - & - & 5.7 & 10.8 \\
        data2vec \cite{baevski2022data2vec}  & 4.0 & 8.5 & 4.6 & 9.1 \\
        \hline
        MS-HuBERT  & 5.6 & 10.9 & 5.9 & 11.3 \\
        - Swap & 5.9 & 11.6 & 6.2 & 12.3 \\
        - Multi & 5.8 & 11.9 & 6.1 & 12.0 \\

        \hline
        \hline
        \multicolumn{5}{c}{\multirow{1}{*}{\textbf{10hr}}} \\    

        \hline
        wav2vec 2.0  & 3.8 & 9.1 & 4.3 & 9.5 \\
        HuBERT   & 3.9 & 9.0 & 4.3 & 9.4 \\
        WavLM  & - & - & 4.3 & 9.2 \\
        data2vec  & 3.3 & 7.5 & 3.9 & 8.1 \\
        \hline
        MS-HuBERT  & 3.6 & 8.5 & 4.1 & 8.8 \\
        - Swap  & 3.8 & 8.6 & 4.1 & 9.2 \\
        - Multi & 3.8 & 9.3 & 4.3 & 9.5 \\

        \hline
        \hline
        \multicolumn{5}{c}{\multirow{1}{*}{\textbf{100hr}}} \\    
        \hline
        
        wav2vec 2.0   & 2.7 & 7.9 & 3.4 & 8.0 \\
        HuBERT   & 2.7 & 7.8 & 3.4 & 8.1 \\
        WavLM  & - & - & 3.4 & 7.7 \\
        data2vec  & 2.2 & 6.4 & 2.8 & 6.8 \\
        \hline
        MS-HuBERT & 2.4 & 7.1 & 3.0 & 7.2 \\
        - Swap  & 2.6 & 6.9 & 3.1 & 7.4 \\
        - Multi & 2.7 & 7.9 & 3.2 & 7.9 \\

        \hline
        \hline
        \multicolumn{5}{c}{\multirow{1}{*}{\textbf{960hr}}} \\    

        \hline
        wav2vec 2.0   & 2.0 & 5.9 & 2.6 & 6.1 \\
        data2vec   & - & - & - & 5.5 \\
        \hline
        MS-HuBERT  & 1.8 & 5.1 & 2.4 & 5.5 \\

        \hline
          \end{tabular}}
\label{table:ASR finetuning}
\end{table}

\renewcommand{\arraystretch}{1.2}
\begin{table}[t]
\caption{SUPERB fine-tuning results. P and F stand for pre-training and fine-tuning respectively. 6-11 means using layers only 6,7,8,9,10,11. For more detailed results using different layers see Section \ref{subsec:evallayersingle}.}
\centering
    \scalebox{0.85}{
    \begin{tabular}{l|c|c|c}
    \hline
        Method & P/F& PR & ASR  \\
        \hline
        \hline
        wav2vec 2.0 \cite{baevski2020wav2vec2.0} & P & 5.74 & 6.43 \\
        \hline
        \hline
        HuBERT \cite{hsu2021hubert} & P & 5.41 & 6.42   \\
        HuBERT + Spin256 \cite{chang2023selfspin} & P + F & 4.39 & 6.34  \\ 
        HuBERT + LASER \cite{meghanani2024laser} & P + F (1/3) & 4.61 &  6.18 \\ 
        \hline
        \hline
        WavLM \cite{chen2022wavlm} & P & 4.84 & 6.21   \\
        WavLM + Spin256 \cite{chang2023selfspin} & P + F  & 4.18 & 6.34   \\
        WavLM + LASER \cite{meghanani2024laser} & P + F (1/3) & 4.28 &  5.92  \\ 
        WavLM Base + (iter 3) (90k) \cite{chen2022wavlm} & P & 3.92 & 5.59  \\
        \hline
        \hline
        ContentVec \cite{qian2022contentvec} & P + F & 4.9 & 5.7 \\
        \hline
        \hline
        data2vec \cite{baevski2022data2vec} & P & 4.69 & \textbf{4.94}  \\
        \hline
        \hline
        MR-HuBERT \cite{shi2023MRHuBERT} (iter 3) & P & 4.16 & 5.76 \\
        \hline
        \hline
        MS-HuBERT & P & 4.42 & 5.60  \\
        -Swap & P & 4.37 & 5.68    \\
        -Multi & P & 5.0 & 6.48    \\
        MS-HuBERT (iter 3) & P & 4.17 & 5.32 \\
        MS-HuBERT (iter 3) (6 - 11) & P & \textbf{4.05} & 5.25 \\

        
        \hline
        
        \end{tabular}}
\label{table:superbfinetuning}
\end{table}

\subsection{Main Results: Supervised Fine-tuning and Inference}

Table \ref{table:ASR finetuning} presents the outcomes on the Librispeech ASR benchmark, where MS-HuBERT is compared with two similar approaches, HuBERT and WavLM. It is evident that MS-HuBERT yields superior results. The margin of improvement increase and the size of dataset used for fine-tuning has a direct proportionality. This is a desired property of any training framework i.e., as the dataset increase the performance should increase.  

Notably, upon the removal of the Swap concept, we observed a degradation in performance, particularly in low-resource settings. This proves that the Swap method does indeed contribute positively to the performance gains.

\subsection{MS-HuBERT as a Feature Extractor}

To study the information encoded/learnt at different layers of the MS-HuBERT model and how it compares to the original HuBERT, we conduct two experiments: (i) SUPERB benchmark \cite{yang2021superb} and (ii) canonical correlation analysis (CCA) similarity with word labels \cite{pasad2023comparative, pasad2021layercpc}.

\noindent \textbf{SUPERB Benchmark}: 
The SUPERB benchmark is designed to evaluate the efficacy of a pre-trained model without fine-tuning i.e., using the frozen encoder as a feature extractor. Specifically, a linear weighted sum of the output embeddings of all the encoder layers serves as a feature for solving any particular downstream task. In our study, we aim to assess the quality of 2nd encoder embeddings, from the MS-HuBERT, for tackling the speech recognition task. Thus, we employ the ASR and phoneme-recognition (PR) tasks within the SUPERB benchmark. The evaluation is on the clean split of the ASR Librispeech benchmark. The results are reported in Table \ref{table:superbfinetuning}. Clearly MS-HuBERT surpasses HuBERT and similar models by a significant margin. This shows the model's capability in encoding information crucial in solving the ASR and PR task. Except on the ASR task using data2vec. 

Based on the above comparison, we hypothesize, that MPL using pseudo labels generated from k-means is better suited for PR task than ASR. The reason might be the clustering algorithm used itself. 

\begin{figure}[ht]
\centering
{
\includegraphics[width=1.0\columnwidth]{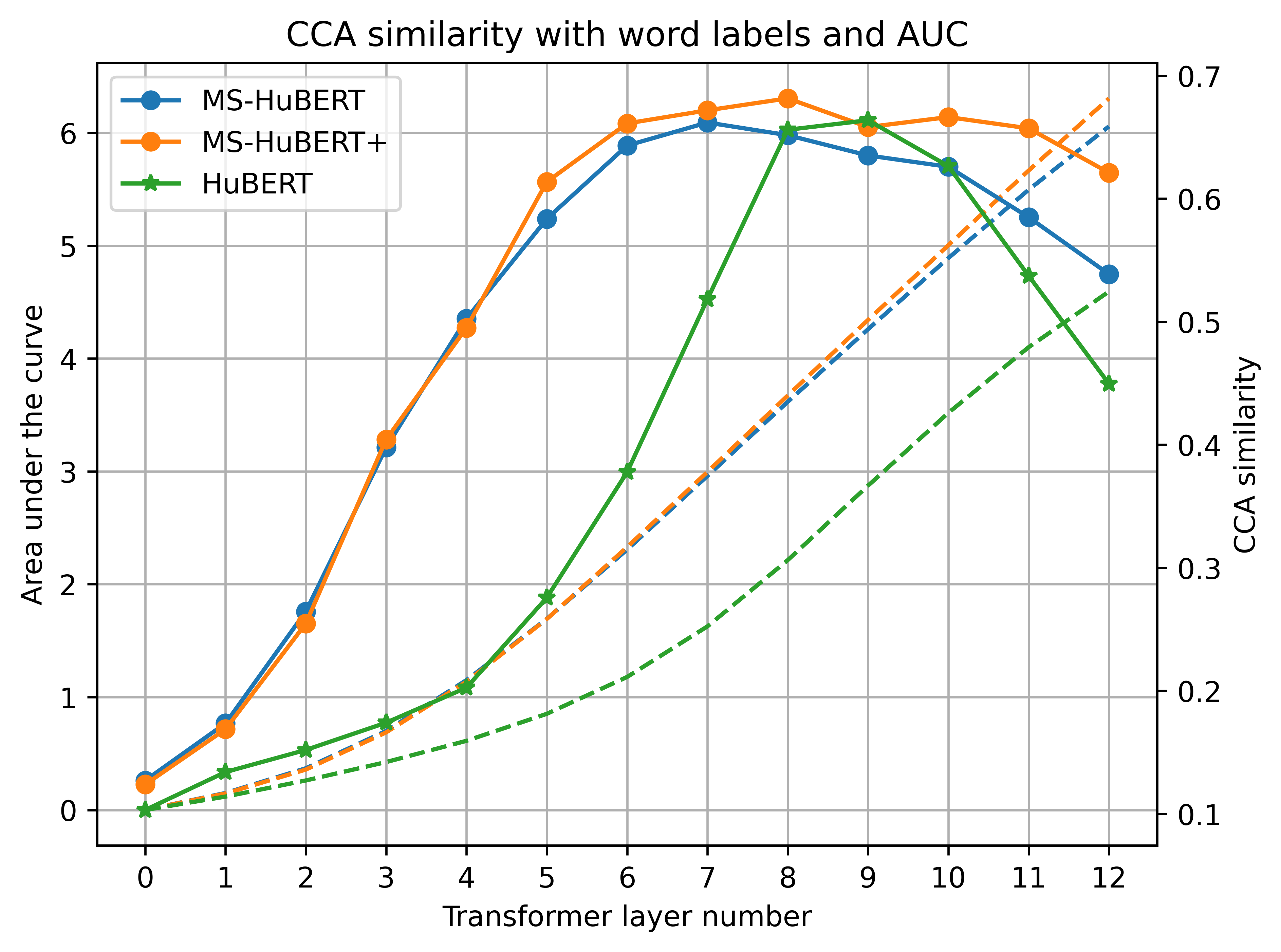}
\caption{Solid lines show the CCA similarity with the word labels. Dotted lines show the AUC area under the curv for different models. }
\label{fig:swaplayerwiseanalysis}
}
\end{figure}

\begin{figure}[ht]
\centering
{
\includegraphics[width=1.0\columnwidth]{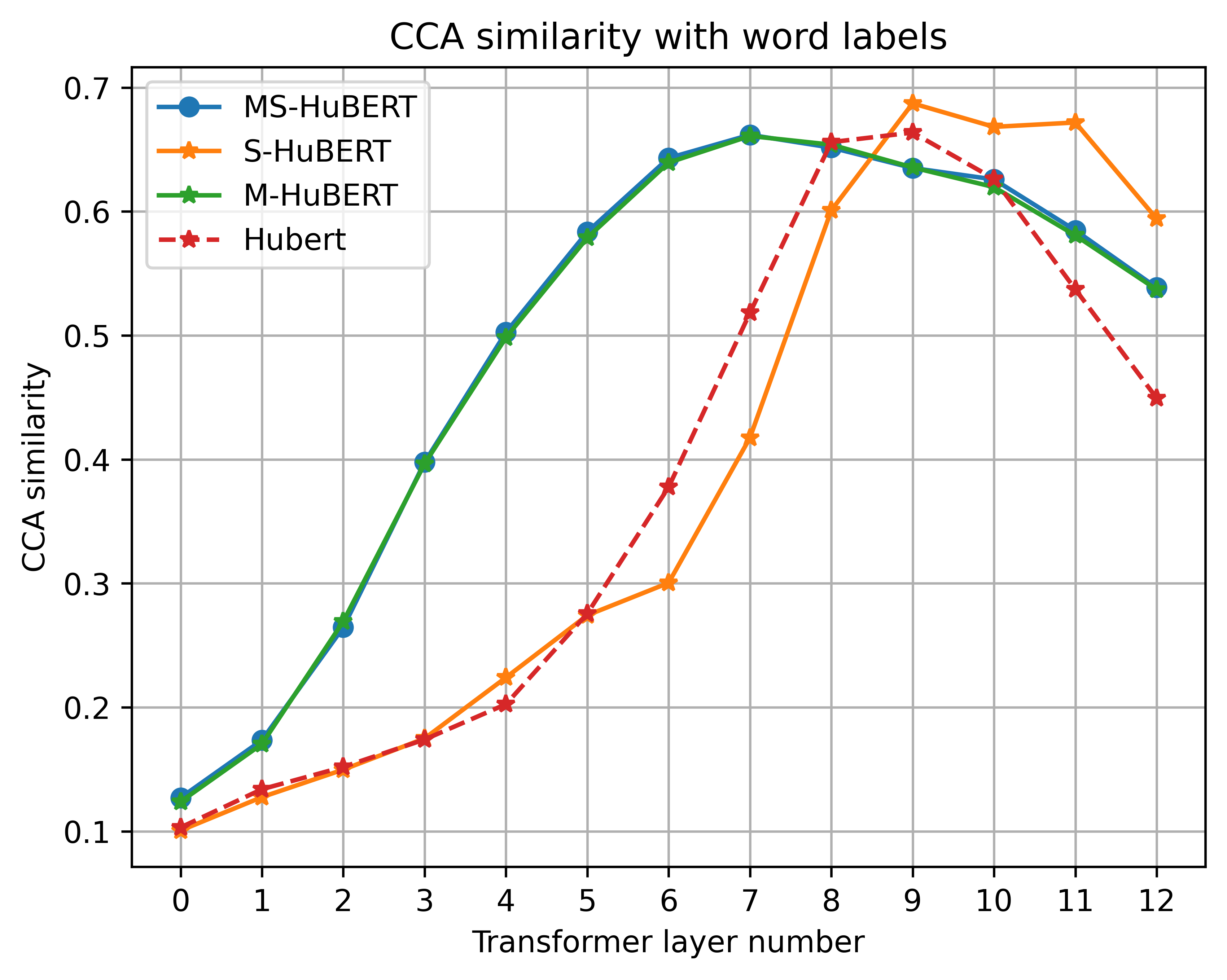}
\caption{CCA similarity with the word labels for MS-HuBERT and its variants. The S-HuBERT curve is similar to WavLM.}
\label{fig:iter2ccaablation}
}
\end{figure}

\noindent \textbf{CCA Similarity with Word Labels}: Following the layer-wise analysis conducted by Pasad et al. \cite{pasad2023comparative, pasad2021layercpc}, we use a modified version of canonical correlation analysis (CCA). Specifically, a projection-weighted CCA (PWCCA) \cite{morcos2018insights}.
The plots are shown in Figure \ref{fig:swaplayerwiseanalysis}. It is clear that MS-HuBERT significantly enhances the performance of word-level information across the transformer encoder layers. Additionally, we compute the area under the curve (AUC) and observe that it consistently surpasses that of Hubert. This increases the model capacity utilization compared to HuBERT. We also plot the M-HuBERT (MS-HuBERT - Swap) to study the effect the of Swap method. We found that combining Swap with the Multicluster loss slightly increases the AUC compared to not using it as shown in Figure \ref{fig:iter2ccaablation}.

\renewcommand{\arraystretch}{1.2}
\begin{table}[ht]
\caption{3rd iteration results using a
4-gram language model. WavLM + is model trained on 90k hours of dataset with data augmentation and 1 million steps.}
\centering
    \scalebox{0.9}{
    \begin{tabular}{l|c|c}
    \hline
    
        Method & test-clean & test-other  \\
        \hline
        \hline
        \multicolumn{3}{c}{\multirow{1}{*}{\textbf{1hr}}} \\    
        \hline
        
        WavLM Base + (iter 3) (90k) \cite{chen2022wavlm}  & 5.4 & 9.8 \\
        MR-HuBERT \cite{shi2023MRHuBERT} (iter 3) & 7.41 & 12.14 \\
        \hline
        Ms-HuBERT  & 5.9 & 11.3 \\
        Ms-HuBERT (iter 3) & 5.4 & 10.8 \\
    
        \hline
        \hline
        \multicolumn{3}{c}{\multirow{1}{*}{\textbf{10hr}}} \\    

        \hline
        WavLM Base + (iter 3) (90k)  & 4.2 & 8.8 \\
        MR-HuBERT \cite{shi2023MRHuBERT} (iter 3) & 4.91 & 8.33 \\
        \hline
        Ms-HuBERT  & 4.1 & 8.8 \\
        Ms-HuBERT (iter 3)  & 3.9 & 8.4 \\
        
        \hline
        \hline
        \multicolumn{3}{c}{\multirow{1}{*}{\textbf{100hr}}} \\

        \hline
        WavLM Base + (iter 3) (90k)   & 2.9 & 6.8 \\
        MR-HuBERT \cite{shi2023MRHuBERT} (iter 3) & 3.57 & 6.81 \\
        \hline
        Ms-HuBERT  & 3.0 & 7.1 \\
        Ms-HuBERT (iter 3)  & 3.0 & 7.0\\
        
        \hline
        
        \end{tabular}}
\label{table:wavlm comparision}
\end{table}

\renewcommand{\arraystretch}{1.2}
\begin{table}[ht]
\caption{ No LM. on the 100hr subset. Encoder-fixed. Encoder-decoder models comparison.}
\centering
    \scalebox{1}{
    \begin{tabular}{l|c|c}
    \hline
    
        Method & test-clean & test-other  \\   
        \hline
        SpeechT5 \cite{ao2021speecht5} & 4.4 & 10.4 \\
        speech2c  \cite{gao2021prespeech2c} & 5.0 & 11.9 \\
        \hline
        MS-HuBERT (iter 3) & 4.2 & 10.2 \\
        \hline
        
        \end{tabular}}
\label{table:speech2c}
\end{table}

\subsection{3rd Iteration Models}
We compare 3rd iteration MS-HuBERT (iter 3) to the 3rd iteration WavLM base+ which is trained on 960 and 94,000 hours of dataset respectively. WavLM is trained for 1 million steps. 3rd iteration MS-HuBERT is trained using the six set pseudo labels generated from the 7th layer of 2nd iteration MS-HuBERT. As shown in Table \ref{table:wavlm comparision}, MS-HuBERT (iter 3) achieves comparable performance to WavLM base +, even though pre-trained using 100 times less data. Which again shows that MS-HuBERT utilizes the model capacity most effectively. 

Table \ref{table:speech2c} presents our results for an encoder-decoder framework inspired by Speech2C \cite{gao2021prespeech2c}. The encoder remains fixed, while only the decoder is trained using six hierarchical clusters, each applied sequentially across the six decoder layers, with fewer clusters assigned to the initial layer. Notably, MS-HuBERT achieves superior performance.

\renewcommand{\arraystretch}{1.2}
\begin{table}[t]
\caption{Evaluation of Individual Layers on the SUPERB Benchmark. PR and ASR tasks are trained for 20\% of the total steps, with results in circular brackets indicating training for 100\% of the total steps. Results in rectangular brackets are for HuBERT \cite{yang2024largelayerhubert}.}
\centering
    \scalebox{1}{
    \begin{tabular}{c|c|c}
    \hline

    \textbf{Layer no/s} & \textbf{PR} & \textbf{ASR}  \\
    \hline
    6 - 11 & (4.05) & (5.25) \\
    \hline
    5 & 8.41 [21.05] & 10.52 \\
    \hline
    6 & 6.56 & 8.75 \\
    \hline
    7 & 5.63 [11.23] & 7.24 \\
    \hline
    8 & 5.01 & 6.24 \\
    \hline
    9 & 4.92 [7.80] & 5.62 \\
    \hline
    10 & 4.41 (4.36) & 5.21 \\
    \hline
    11 & 4.60 [6.10] & 5.04 (4.85) \\
    \hline
    12 & 4.82 & 5.17 \\
    \hline

        \end{tabular}}
\label{table:singlelayersuperb}
\end{table}

\subsection{Evaluation of Individual Layers on the SUPERB Benchmark}
\label{subsec:evallayersingle}
\begin{figure}[t]
\includegraphics[width=1\columnwidth, trim=0 0 0 3, clip]{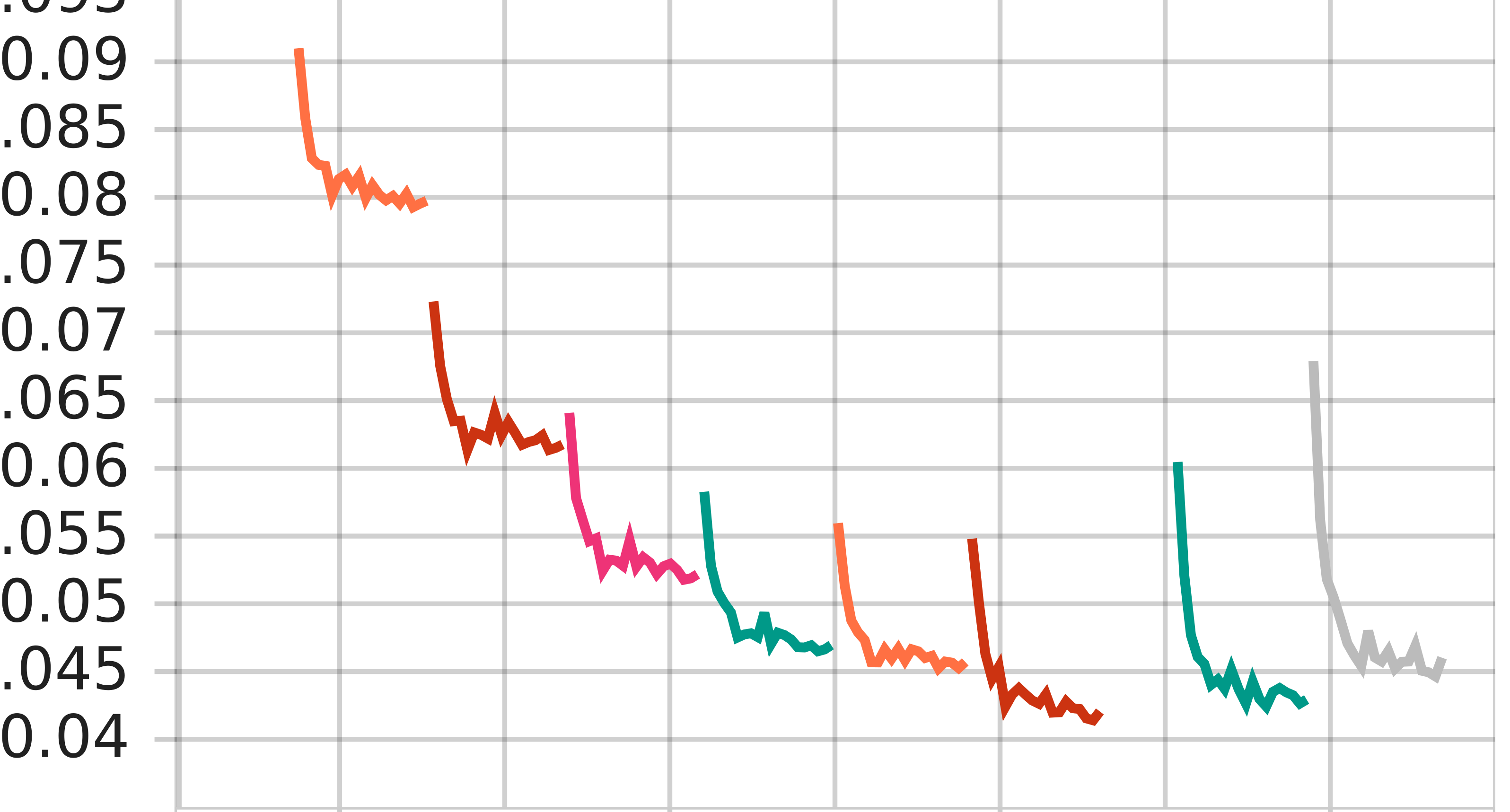}
\caption{PER values are plotted with the x-axis representing layers 5 to 12, ordered from left to right.}\label{fig:singlelayersuperbper}
\end{figure}

\begin{figure}[t]
\includegraphics[width=1\columnwidth, trim=0 23 0 4, clip]{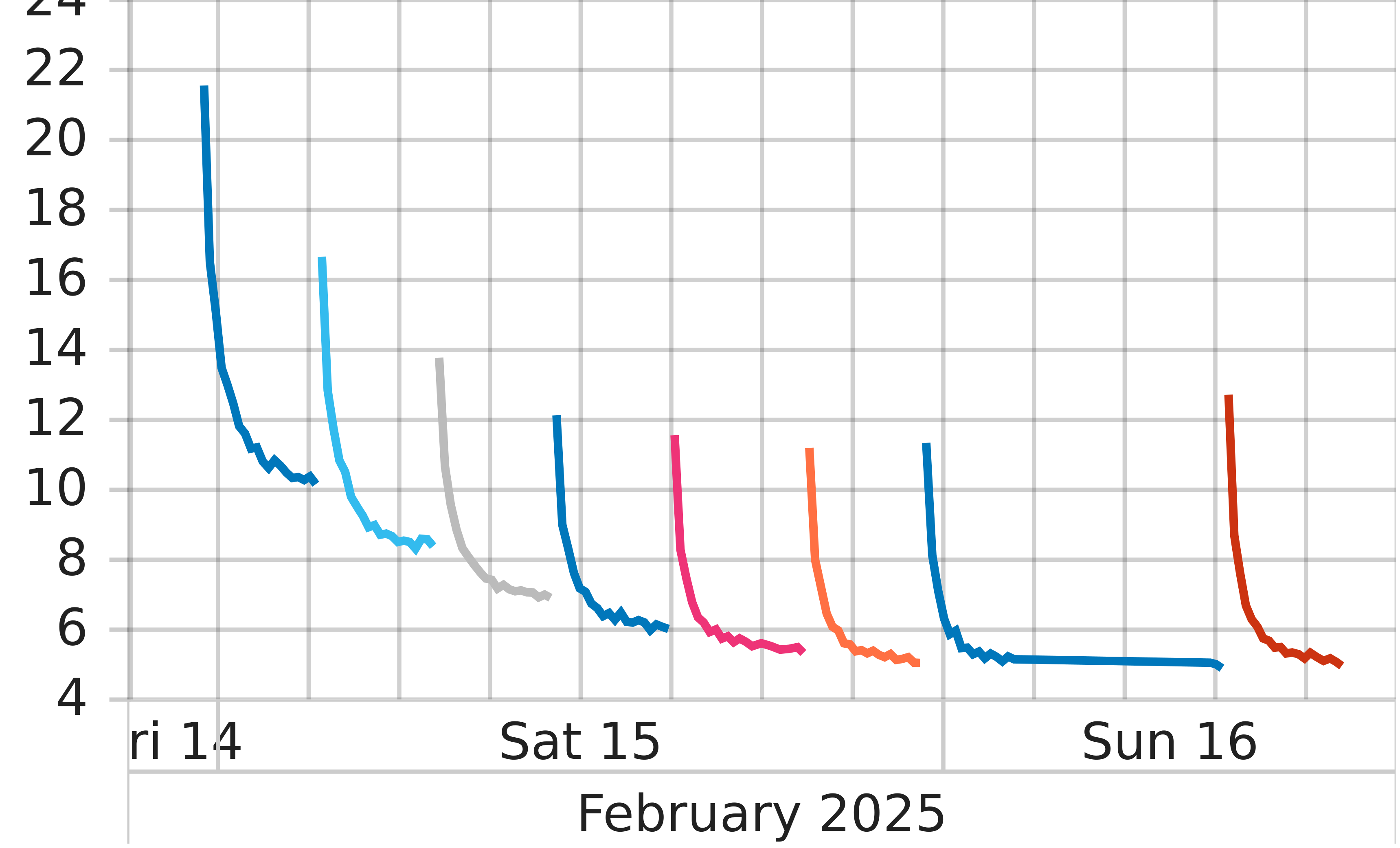}
\caption{WER values on the y-axis should be divided by 100 to obtain the final WER. The x-axis represents layers from 5 to 12, ordered from left to right.}\label{fig:singlelayersuperbwer}
\end{figure}

Rather than computing a weighted average across all layers, we analyze the performance of using a single layer with a high CCA score, specifically from layer 5 onward. The results are presented in Table \ref{table:singlelayersuperb} and Figures \ref{fig:singlelayersuperbper} and \ref{fig:singlelayersuperbwer}.
In summary, We found that a weighted average yields the best performance for the PR task, while using a specific layer works best for the ASR task. Overall depth is important for learning better representations. 

Lastly, using the 8th layer of MS-HuBERT improves over HuBERT, resulting in 33\% less compute of transformer encoder.

\section{Discussion}
\label{section:discussion}
In comparison to data2vec \cite{baevski2022data2vec}, our performance on the ASR Librispeech benchmark, as illustrated in Table \ref{table:ASR finetuning}, still falls short, particularly evident in low resource scenarios. This difference may stem from the inherent nature of the MLM pre-text task utilizing discrete tokens. For instance, WER metric for HuBERT and WavLM in a 1-hour setting lag behind even wav2vec 2.0. However, as the fine-tuning dataset increases, the performance gap diminishes. When leveraging the entire 960 hours of the Librispeech dataset, our performance matches that of data2vec. On the SUPERB benchmark, for the PR task, MS-HuBERT outperforms data2vec and is comparable in the context of ASR.

Given MS-HuBERT is trained using the pseudo labels generated from the the first iteration HuBERT, there could be a scope for performance improvements as is evident from the MS-HuBERT (iter 3). Or the number of layers used in between the Multicluster MPL may restrict the capacity to learn higher level embeddings. However, larger models may alleviate this limitation by offering more layers in between the MPL calculation.



\section{Conclusion and Future Work}
Our results highlight the potential of MS-HuBERT in bridging the performance gap between HuBERT and data2vec on the ASR Librispeech benchmark and content based tasks, ASR and PR, on the SUPERB benchmark. 
MS-HuBERT is aimed at mitigating the pre-training and inference mismatch in masked language modeling for learning. 
Building upon the HuBERT framework, MS-HuBERT incorporates two key modifications: the Swap method, enabling full context access during pre-training, and the Multicluster loss approach for more effective training. 
Through empirical evaluation on the ASR Librispeech benchmark, MS-HuBERT demonstrates significant performance improvements over the original HuBERT model, achieving state-of-the-art results and matching the performance of data2vec in high-resource settings. 
Future research could explore further enhancements to the MS-HuBERT methodology to avoid iterative pre-training or improving the quality pseudo labels altogether. Lastly, scaling the model size is also an open question. 

\section{Limitations}
Complexity and Computational Cost: MS-HuBERT introduces additional complexity to the pre-training process, particularly with the incorporation of the Swap method and a little overhead when using Multicluster MPL. This increased complexity may result in higher computational costs during pre-training only. But during inference, MS-HuBERT and HuBERT has the same complexity as they share exactly the same architecture.

\bibliographystyle{IEEEtran}
\bibliography{main}

\appendix
\section{appendix}
\subsection{CCA}
\begin{figure}[ht]
\includegraphics[width=1\columnwidth]{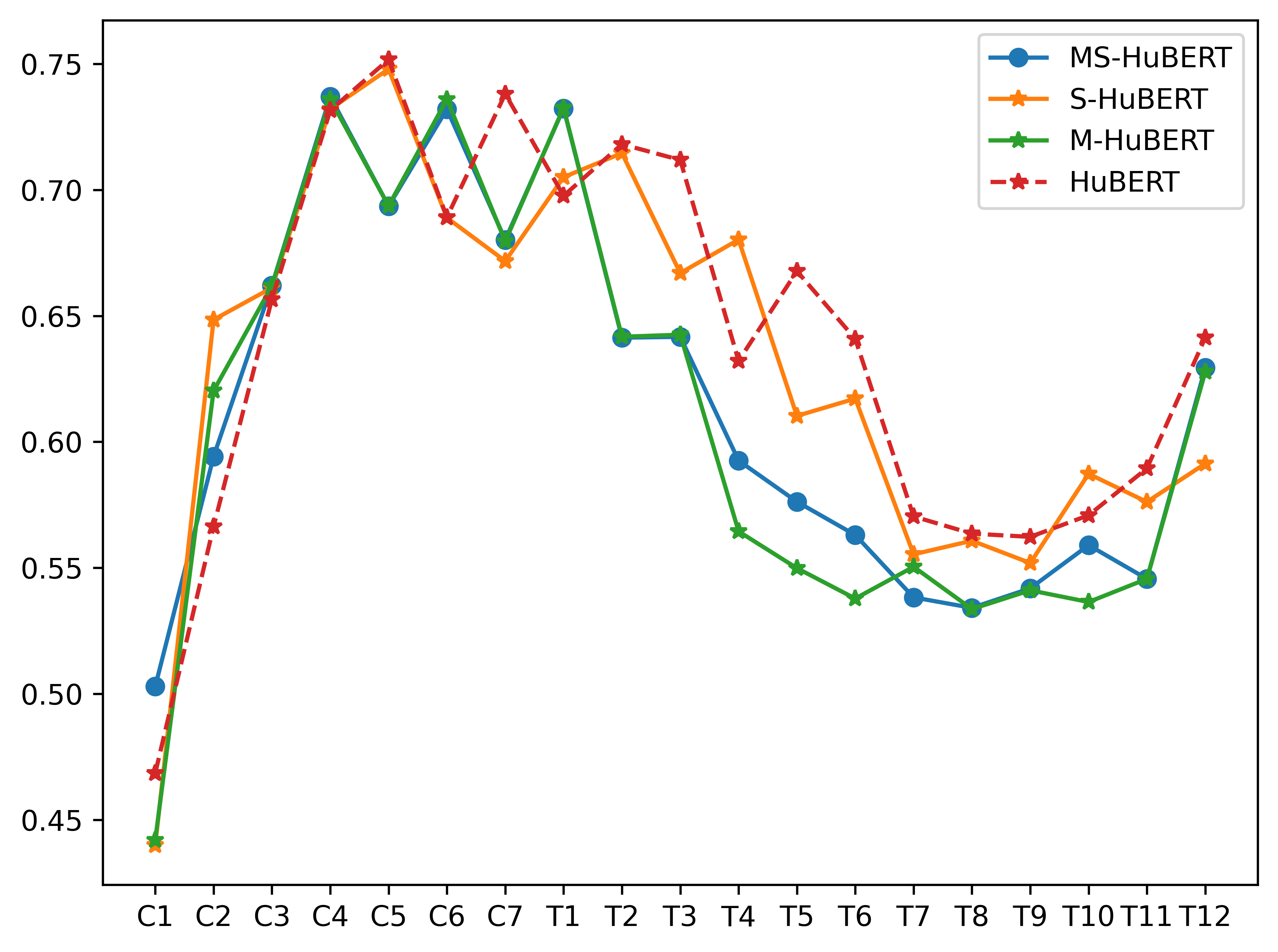}
\caption{CCA similarity with the mel}\label{fig:MS-M-HuBERT_mel}
\end{figure}

\begin{figure}[ht]
\includegraphics[width=1\columnwidth]{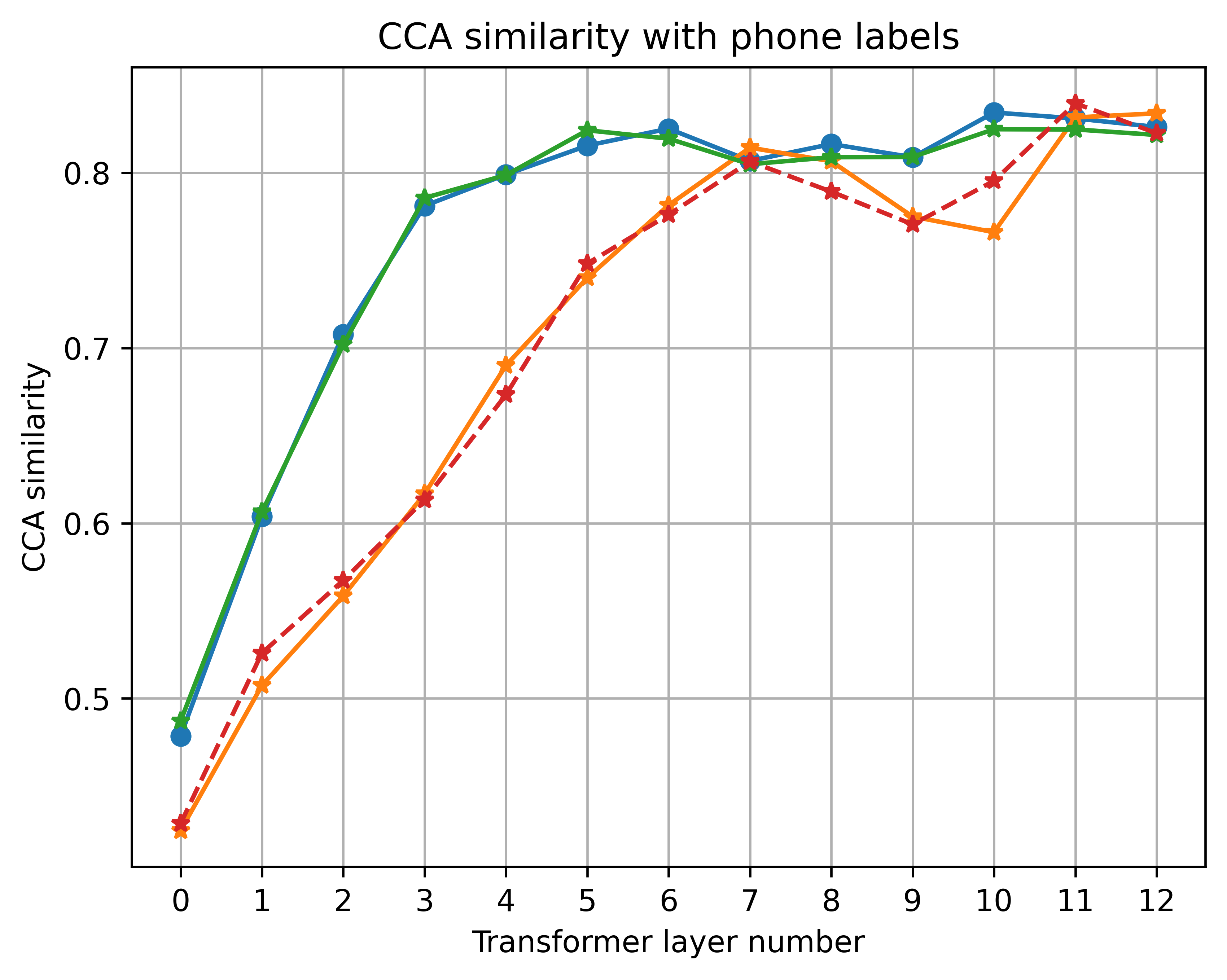}
\caption{CCA similarity with the phone}\label{fig:MS-M-HuBERT_phone}
\end{figure}

\begin{figure}[ht]
\includegraphics[width=1\columnwidth]{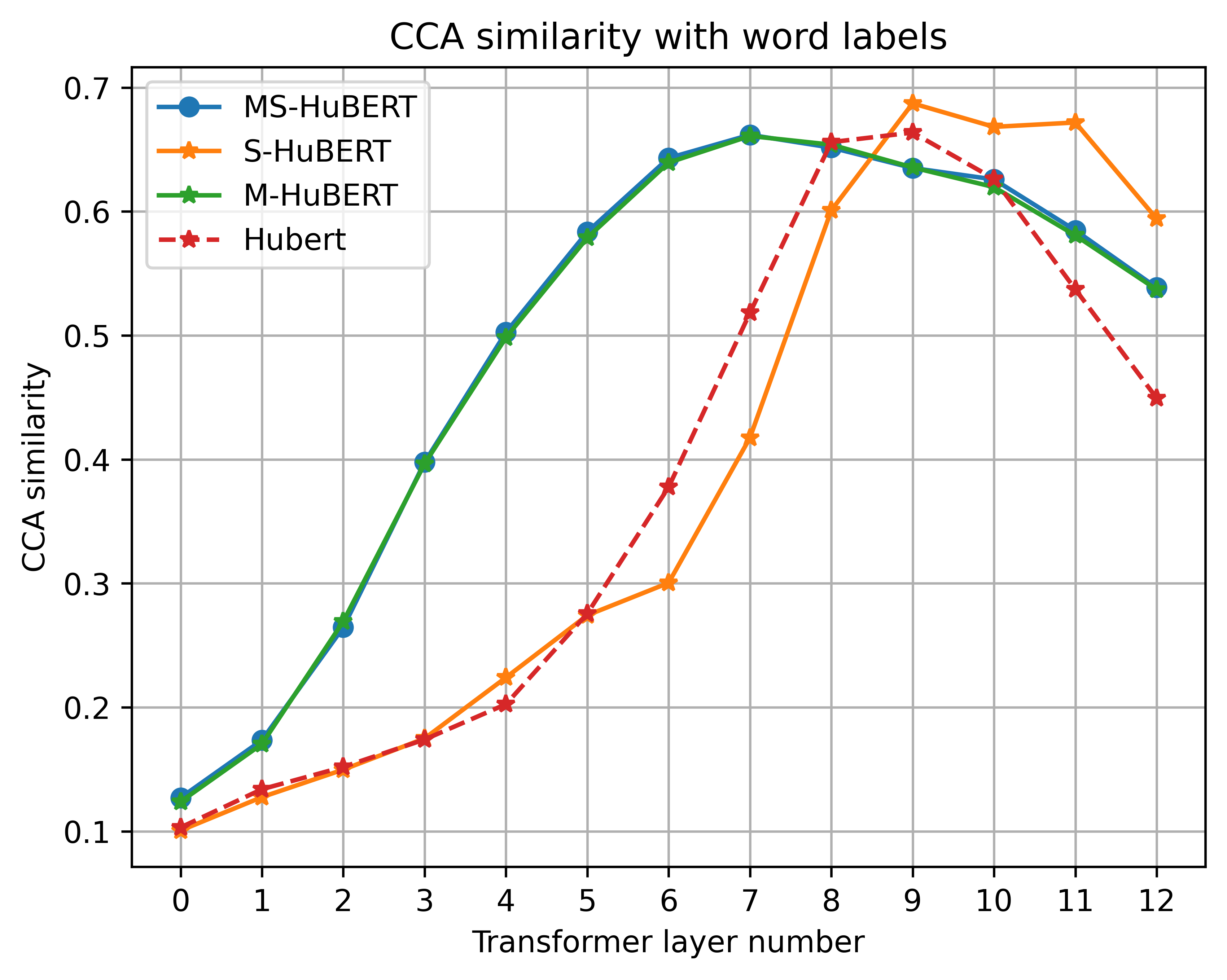}
\caption{CCA similarity with the word}\label{fig:MS-M-HuBERT_word}
\end{figure}

Figure \ref{fig:MS-M-HuBERT_mel}, \ref{fig:MS-M-HuBERT_phone} and, \ref{fig:MS-M-HuBERT_word} shows the plots for CCA similarity for different settings similar to the original work in \cite{pasad2021layercpc, pasad2023comparative}.

\subsection{MS-HuBERT components}
Table \ref{table:ASR ablation} shows the ablation results for different components of MS-HUBERT (iter 2).


\renewcommand{\arraystretch}{1.0}
\begin{table}[ht]
\caption{ASR ablation}
\centering
    \scalebox{1.0}{
    \begin{tabular}{|lccc|}
    \hline
    
      Method & LM & test-lean  & test-other  \\
        \hline
        \hline
        \multicolumn{4}{c}{\multirow{1}{*}{\textbf{1hr}}} \\   
        \hline

        \hline
        MS-HuBERT & None & 18.4 & 24.8 \\
        M-HuBERT & None & 19.5 & 26.3 \\
        S-HuBERT & None   & 21.3 & 28.2 \\

        \hline

        MS-HuBERT & 4-gram & 5.9 & 11.3 \\
       M-HuBERT & 4-gram & 6.2 & 12.3 \\
       S-HuBERT & 4-gram  & 6.1 & 12.2 \\

        \hline
        \hline
        \multicolumn{4}{c}{\multirow{1}{*}{\textbf{10hr}}} \\    
        \hline

        \hline
        MS-HuBERT & None & 8.2 & 14.7 \\
        M-HuBERT & None & 8.5 & 14.9 \\
        S-HuBERT & None & 9.7 & 17.0 \\

        \hline
        MS-HuBERT & 4-gram &  4.1 & 8.8 \\
        M-HuBERT & 4-gram & 4.1 & 9.2 \\
        S-HuBERT & 4-gram & 4.3 & 9.5 \\

        \hline
        \hline
        \multicolumn{4}{c}{\multirow{1}{*}{\textbf{100hr}}} \\    
        \hline

        \hline
        MS-HuBERT & None & 4.6 & 10.6 \\
        M-HuBERT & None & 4.8 & 10.9 \\
        S-HuBERT & None & 5.5 & 12.5 \\

        \hline
        MS-HuBERT & 4-gram & 3.0 & 7.2 \\
        M-HuBERT & 4-gram & 3.1 & 7.4 \\
        S-HuBERT & 4-gram & 3.2 & 7.9 \\

        \hline

        \end{tabular}}
\label{table:ASR ablation}
\end{table}

\end{document}